\documentclass[sigconf]{acmart}
\usepackage{amsthm,color}
\usepackage{multirow, tabularx}
\usepackage{subcaption}
\usepackage{graphicx}

\AtBeginDocument{%
  \providecommand\BibTeX{{%
    \normalfont B\kern-0.5em{\scshape i\kern-0.25em b}\kern-0.8em\TeX}}}

\setcopyright{iw3c2w3}
\begin{document}

\copyrightyear{2021}
\acmYear{2021} 
\acmConference[WWW '21]{Proceedings of the Web Conference 2021}{April 19--23, 2021}{Ljubljana, Slovenia} 
\acmBooktitle{Proceedings of the Web Conference 2021 (WWW '21), April 19--23, 2021, Ljubljana, Slovenia}
\acmPrice{}
\acmDOI{10.1145/3442381.3449971}
\acmISBN{978-1-4503-8312-7/21/04}

\title{HDMI: High-order Deep Multiplex Infomax}


\author{Baoyu Jing}
\affiliation{%
  baoyuj2@illinois.edu\\
  \institution{University of Illinois at Urbana-Champaign}
  \state{IL}
  \country{USA}
}

\author{Chanyoung Park}
\email{cy.park@kaist.ac.kr}
\affiliation{%
  \institution{Department of Industrial and Systems Engineering, KAIST}
  \state{Daejeon}
  \country{Republic of Korea}
}

\author{Hanghang Tong}
\email{htong@illinois.edu}
\affiliation{%
  \institution{University of Illinois at Urbana-Champaign}
  \state{IL}
  \country{USA}
}


\begin{abstract}
Networks have been widely used to represent the relations between objects such as academic networks and social networks, and learning embedding for networks has thus garnered plenty of research attention.
Self-supervised network representation learning aims at extracting node embedding without external supervision.
Recently, maximizing the mutual information between the local node embedding and the global summary (e.g. Deep Graph Infomax, or DGI for short) has shown promising results on many downstream tasks such as node classification.
However, there are two major limitations of DGI.
Firstly, DGI merely considers the  \textit{extrinsic} supervision signal (i.e., the mutual information between node embedding and global summary) while ignores the \textit{intrinsic} signal (i.e., the mutual dependence between node embedding and node attributes). 
Secondly, nodes in a real-world network are usually connected by multiple edges with different relations, while DGI does not fully explore the various relations among nodes.
To address the above-mentioned problems, we propose a novel framework, called High-order Deep Multiplex Infomax (HDMI), for learning node embedding on multiplex networks in a self-supervised way.
To be more specific, we first design a joint supervision signal containing both extrinsic and intrinsic mutual information by high-order mutual information, and we propose a High-order Deep Infomax (HDI) to optimize the proposed supervision signal. 
Then we propose an attention based fusion module to combine node embedding from different layers of the multiplex network.
Finally, we evaluate the proposed HDMI on various downstream tasks such as unsupervised clustering and supervised classification. 
The experimental results show that HDMI achieves state-of-the-art performance on these tasks.

\end{abstract}

\ccsdesc[500]{Computing methodologies~Neural networks}
\ccsdesc[300]{Computing methodologies~Learning latent representations}
\ccsdesc[500]{Mathematics of computing~Information theory}
\ccsdesc[500]{Mathematics of computing~Graph algorithms}

\keywords{Network Representation Learning, Multiplex Networks, High-order Mutual Information}

\maketitle

\section{Introduction}
Network or graph structures have been widely used to represent the relations among objects, such as the citation relation among papers, the co-author relation among researchers,
as well as the friendship relation among people.
Mining and discovering useful knowledge from networks has been an active research area, within which network representation learning has garnered substantial research attention and it has been demonstrated to be effective for various tasks on networks \cite{cui2018survey, zhou2018graph, zhang2019graph}, such as node classification, node clustering and link prediction.

Self-supervised network representation learning aims at extracting node embedding without introducing external supervision signals. 
The dominant strategy of the self-supervised network representation learning is to design a signal based on the {\em proximity} of the nodes in the network, such that the node embedding will retain the proximity.
For example, given a node, DeepWalk \cite{perozzi2014deepwalk} and node2vec \cite{grover2016node2vec} maximize the probabilities of its neighbors sampled by random walks.
LINE \cite{tang2015line} maximizes the probabilities between a node and its first or second-order proximate neighbors.
Albeit the effectiveness of these methods, the proximity based supervision signals only capture the \textit{local} characteristic of networks.
With the introduction of the powerful network encoders, such as Graph Convolutional Network (GCN) \cite{kipf2016semi} which can naturally capture such local proximity based on graph convolution \cite{bruna2013spectral}, the improvement brought by the traditional proximity based supervision signals is limited \cite{velivckovic2018deep}.
To further improve the quality of node embedding, Deep Graph Infomax (DGI) \cite{velivckovic2018deep} uses GCN as the network encoder and trains it by maximizing the mutual information between the node embedding and the \textit{global} summary of the network.
However, there are two major limitations of DGI.

One limitation of DGI is that it focuses on the \textit{extrinsic} supervision signal (i.e., whether the node embedding and the global summary come from the same network), and it does not fully explore the \textit{intrinsic} signal (i.e., the mutual dependence between the embedding vector and the attribute vector of a node).
Node attributes themselves often contain discriminative information about the nodes, and capturing the mutual dependence between them helps the embedding obtain more discriminative information and thus discover the potential relations among nodes. 
For example, in the citation network, papers always contain textual attributes such as abstracts and keywords.
Two papers focusing on different sub-areas (e.g. dynamic graph algorithms and sub-graph mining) might not have a direct citation relation, but they might share the same or similar keywords (e.g. social network).
In such a case, they are likely to belong to the same area, e.g., social analysis or web mining.
The extrinsic signal only models the global property of the citation network and ignores the intrinsic mutual dependence, and thus it might be unable to discover the hidden relation between the two papers.
The existing dominant strategy for capturing the mutual dependence between embedding and attributes is using reconstruction error of attributes given node embedding \cite{zhang2018anrl,meng2019co,gao2018deep}.
However, as pointed out by \cite{yang2019deep}, there is no substantial connection between the discriminative ability (or quality) of embedding with the reconstruction loss. 
In this paper, we propose to use mutual information between node embedding and node attributes as the intrinsic supervision signal.
Moreover, to further capture the synergy between the extrinsic and the intrinsic signals, we propose to use high-order mutual information among the node embedding, global summary vector, and node attributes, and we propose a novel High-order Deep Infomax (HDI) 
to maximize the high-order mutual information.

\begin{figure}[t]
    \centering
    \includegraphics[width=.5\textwidth]{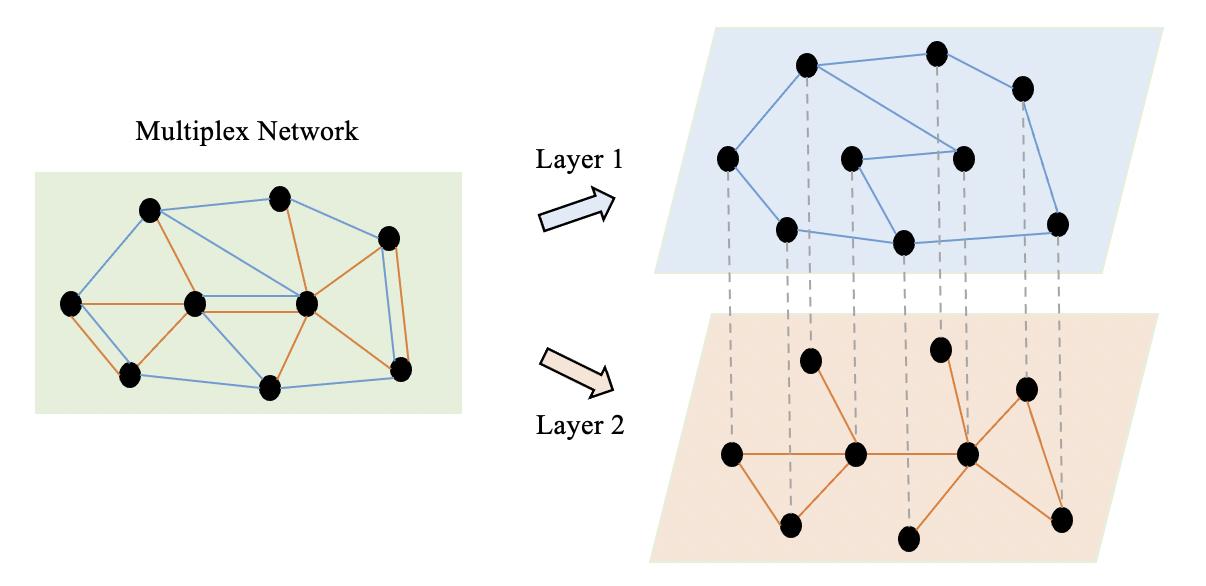}
    \caption{An example of the multiplex network, where different colors represent different types of relations among nodes.
    A multiplex network contains multiple layers of networks, where each layer has one type of relations.}
    \label{fig:example}
\end{figure}

Another limitation of DGI is that it assumes a single type of relation among the nodes in a network, but nodes in a network are usually connected via multiple edges with different relation types.
For example, in an academic network, two papers can be connected via shared authors or citation relation.
In a product network,
products can be linked by relations such as AlsoView, AlsoBought, and BoughtTogether. 
In a social network, two people can be connected by many relations such as direct friendship, the same school, and the same employer.
A network with multiple types of relations is referred to as a multiplex network, and a multiplex network can be decomposed into multiple layers of networks, where each layer only has one type of relation.
Figure \ref{fig:example} 
provides an example of a multiplex network.
The simplest way to learn node embedding for a multiplex network is to first extract node embedding independently from different layers and then use average pooling over the node embedding from different layers to obtain final embedding for each node. 
However, as observed by many recent studies \cite{wang2019heterogeneous, park2020unsupervised,cen2019representation,shi2018mvn2vec,zhang2018scalable}, layers are related with each other and they can mutually help each other for downstream tasks.
For example, in the academic multiplex network mentioned above, the citation network layer and the shared author network layer usually provide different aspects of the papers' subject.
To be more specific, the citation layer usually links a paper (e.g. about the attributed network) with related papers from other areas (e.g. computer vision), while a specific author usually focuses on a specific area (e.g. network representation learning).
To capture such a mutual relation, the attention mechanism \cite{bahdanau2014neural} is mostly adopted to calculate the weights for different layers \cite{wang2019heterogeneous, qu2017attention}.
However, these methods require external supervision (e.g. labels of nodes) to train the attention module. 
Recently, DMGI \cite{park2020unsupervised} proposes a complex combination module, which first uses an attention mechanism to obtain reference node embedding and then uses a consensus regularization to obtain final node embedding.
In this paper, we propose an alternative semantic attention \cite{you2016image} based method as the fusion module to combine node embedding from different layers.
More importantly, different from existing works, the proposed fusion module is trained via the proposed high-order mutual information. 
\newpage
Our main contributions are summarized as follows:
\begin{itemize}
    \item We propose a novel supervision signal based on the high-order mutual information, which combines extrinsic and intrinsic mutual information, for learning network embedding on both attributed networks and attributed multiplex networks.
    \item We introduce a novel High-order Deep Infomax (HDI) to optimize the proposed high-order mutual information based supervision signal. 
    \item We propose an attention based fusion module to combine node embedding from different layers of a multiplex network, which is trained via the high-order mutual information based supervision signal.
    \item We evaluate the proposed methods on a variety of real-world datasets with various evaluation metrics to demonstrate the effectiveness of the proposed methods.
\end{itemize} 

The rest of the paper is organized as follows.
We briefly introduce the attributed multiplex network, mutual information and DGI in Section \ref{sec:prelimiary}.
We introduce the high-order mutual information, as well as the proposed HDI and HDMI in Section \ref{sec:methodology}.
The experimental results are presented in Section \ref{sec:experiments}.
We provide a brief review of the most relevant works in Section \ref{sec:related_works}.
Finally, we conclude the paper in Section \ref{sec:conclusion}.

\section{Preliminaries}\label{sec:prelimiary}
In this section, we first introduce relevant concepts for attributed multiplex networks.
Then we introduce mutual information related concepts as well as Deep Graph Infomax (DGI) \cite{velivckovic2018deep}.
We also summarize the notations used in this paper in Table \ref{tab:notations}.

\subsection{Attributed Multiplex Networks}
An attributed multiplex network (Definition~\ref{def:attr_multiplex}) is comprised of multiple layers of attributed networks (Definition~\ref{def:attr_homo_net}). 

\begin{definition}[Attributed Network]\label{def:attr_homo_net}
An attributed network is represented by $\mathcal{G}(A, F)$, where $A\in\mathbb{R}^{N\times N}$ denotes adjacency matrix and $F\in\mathbb{R}^{N\times d_F}$ denotes the attribute matrix, $N$ and $d_F$ denote the number of nodes and the dimension of attributes respectively.
\end{definition}

\begin{definition}[Attributed Multiplex Network]\label{def:attr_multiplex}
An attributed multiplex network $\mathcal{G_M}=\{\mathcal{G}^{1},\cdots,\mathcal{G}^{R}\}$ is comprised of $R\geq1$ layers of attributed networks, where $\mathcal{G}^{r}(A^{r}, F)$ ($r\in[1,\cdots,R]$, $A^{r}\in\mathbb{R}^{N\times N}$, $F\in\mathbb{R}^{N\times d_F}$) denotes the $r$-th layer.
Note that different layers capture different types of relations between nodes, and all of the layers share the same node attribute matrix.
\end{definition}

\subsection{Mutual Information}
Mutual information measures the mutual dependence between two random variables, which is based on Shannon entropy, and its formal definition is given in Definition \ref{def:mi}. 
In order to measure the mutual dependence between multiple random variables, the concept of high-order/multivariate mutual information is introduced in the information theory \cite{mcgill1954multivariate}.

\begin{definition}[Mutual Information]\label{def:mi}
Given two random variables $X$ and $Y$, the mutual information between them is defined by:
\begin{equation}
    I(X;Y) = H(X) + H(Y) - H(X,Y) = H(X) - H(X|Y)
\end{equation}
where $H(X)$ and $H(X|Y)$ denote entropy and conditional entropy respectively,
$H(X,Y)$ denotes the entropy for the joint distribution of $X$ and $Y$.
\end{definition}

\begin{definition}[High-order Mutual Information \cite{mcgill1954multivariate}]\label{def:hmi}
High-order mutual information is a generalization of mutual information on $N\geq3$ random variables.
Given a set of $N$ random variables ${X_1, \cdots, X_N}$, the high-order mutual information is defined analogous to the definition of mutual information (Definition \ref{def:mi}):
\begin{equation}
    I(X_1;\cdots; X_N) = \sum_{n=1}^N(-1)^{n+1}\sum_{i_1<\cdots<i_n}H(X_{i_1},\cdots, X_{i_n})
\end{equation}
where  $H(X_{i_1},\cdots, X_{i_n})$ denotes the joint entropy for $X_{i_1}, \cdots, X_{i_n}$,
and the summation $\sum_{i_1<\cdots<i_n}H(X_{i_1},\cdots, X_{i_n})$ runs over all of the combinations of random variables ($\{i_1,\cdots,i_n\}\in[1,\cdots,N]$).
\end{definition}
High-order mutual information not only captures the mutual information between each pair of two random variables but also the synergy among multiple random variables.
Equation \eqref{eq:n=3}-\eqref{eq:interaction_information} provide an example when $N=3$.

\subsection{Deep Graph Infomax}\label{sec:dgi}
DGI is a self-supervised learning or an unsupervised learning method for learning node embedding on attributed networks (Definition \ref{def:attr_homo_net}), the main idea behind which is to maximize the mutual information between node embedding $\mathbf{h}$ and the global summary vector $\mathbf{s}$ of the entire network $\mathcal{G}$: $I(\mathbf{h}, \mathbf{s})$.

When maximizing $I(\mathbf{h}, \mathbf{s})$, DGI leverages negative sampling strategy.
Specifically, it first generates a negative network $\Tilde{\mathcal{G}}$ via a corruption function $\Tilde{\mathcal{G}}=\mathcal{C}(\mathcal{G})$. 
Then it uses the same encoder $\mathcal{E}$ (e.g. GCN \cite{kipf2016semi}) to obtain the node embedding for the positive network $\{\mathbf{h}_1, \cdots, \mathbf{h}_N\}$ as well as the embedding for the negative network $\{\Tilde{\mathbf{h}}_1, \cdots, \Tilde{\mathbf{h}}_N\}$. Here, $N$ is the number of nodes in both of the networks.
The summary vector of the positive network $\mathcal{G}$ is obtained by a readout function $\mathbf{s}=\mathcal{R}(\{\mathbf{h}_1, \cdots, \mathbf{h}_N\})$ (e.g. average pooling). 
Finally, given $\mathbf{s}$, a discriminator $\mathcal{D}$ is used to distinguish the node embedding of the positive network $\mathbf{h}_n$ with the one from the negative network $\Tilde{\mathbf{h}}_n$.
For more details, please refer to \cite{velivckovic2018deep}.

Maximizing $I(\mathbf{h}_n, \mathbf{s})$ is equivalent to maximize the objective function given in the following definition:

\begin{definition}[Deep Graph Infomax]\label{def:dgi}
Given a node embedding $\mathbf{h}_n$ and a summary vector $\mathbf{s}$ of the attributed network and a negative node embedding $\Tilde{\mathbf{h}}_n$, the mutual information between $\mathbf{h}_n$ and $\mathbf{s}$ can be maximized based on the following objective function:
\begin{equation}\label{eq:dgi}
    \mathcal{L} = \sup_\Theta{\mathbb{E}[\log\mathcal{D}(\mathbf{h}_n; \mathbf{s})] + \mathbb{E}[\log(1 - \mathcal{D}(\Tilde{\mathbf{h}}_n; \mathbf{s}))]}
\end{equation}
where $\mathcal{D}$ denotes the discriminator aiming at distinguish the negative node embedding $\Tilde{\mathbf{h}}_n$ with the real embedding $\mathbf{h}_n$, $\Theta$ denotes parameters, $\mathbb{E}$ denotes the expectation.
\end{definition}

\begin{table}[t]
    \centering
    \caption{Notations}
    \begin{tabular}{l|l}
        \hline
         Symbols &  Descriptions\\
         \hline
         \hline
         $\mathcal{G}$& attributed network\\
         $\mathcal{G_M}$& attributed multiplex network\\
         $\mathcal{C}$ & corruption function \\
         $\mathcal{R}$ & readout function \\
         $\mathcal{E}$ & encoder function \\
         $\mathcal{D}$ & discriminator\\
         $\mathcal{L}$ & objective function\\
         $A$ & adjacency matrix \\
         $F$ & node attribute matrix \\
         $H$ & node embedding matrix \\
         \hline
         $I(X;Y)$ & mutual information between $X$ and $Y$\\
         $H(X)$ & entropy of the random variable $X$\\
         $H(X|Y)$ & conditional entropy of $X$ given $Y$\\
         \hline
         $\mathbf{s}$ & global summary vector\\
         $\mathbf{h}$ & node embedding vector\\
         $\mathbf{f}$ & node attribute vector\\
         \hline
    \end{tabular}
    \label{tab:notations}
\end{table}

\section{Methodology}\label{sec:methodology}
In this section, we first introduce High-order Deep Infomax (HDI) for optimizing high-order mutual information on attributed networks in Section \ref{sec:hdi} and then extend the proposed HDI to attributed multiplex networks by introducing High-order Deep Multiplex Infomax (HDMI) in Section \ref{sec:hdmi}. 

\begin{figure*}
\centering
\begin{subfigure}[b]{.75\textwidth}
    \includegraphics[width=\textwidth]{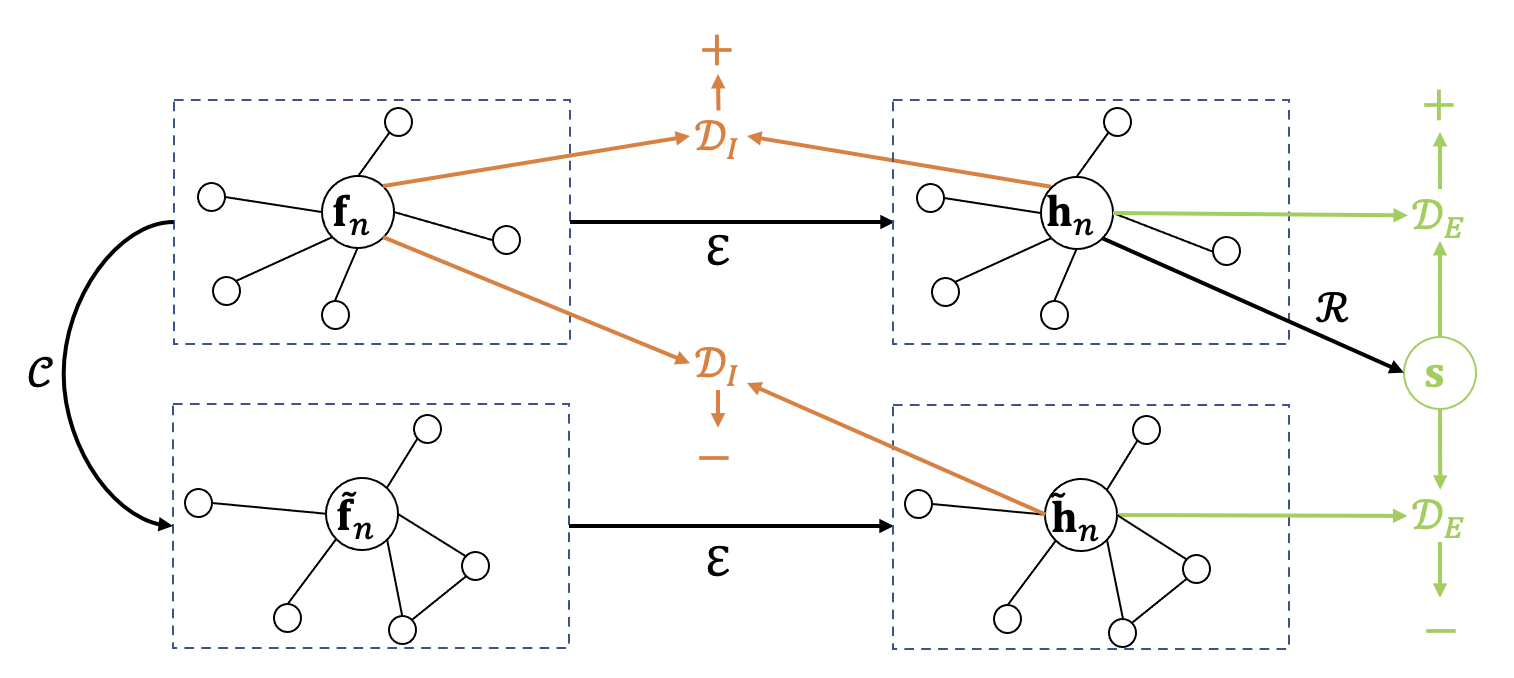}
    \caption{Illustration of the extrinsic and intrinsic supervision.}\label{fig:model_ex_in}
\end{subfigure}
\begin{subfigure}[b]{.7\textwidth}
    \includegraphics[width=\textwidth]{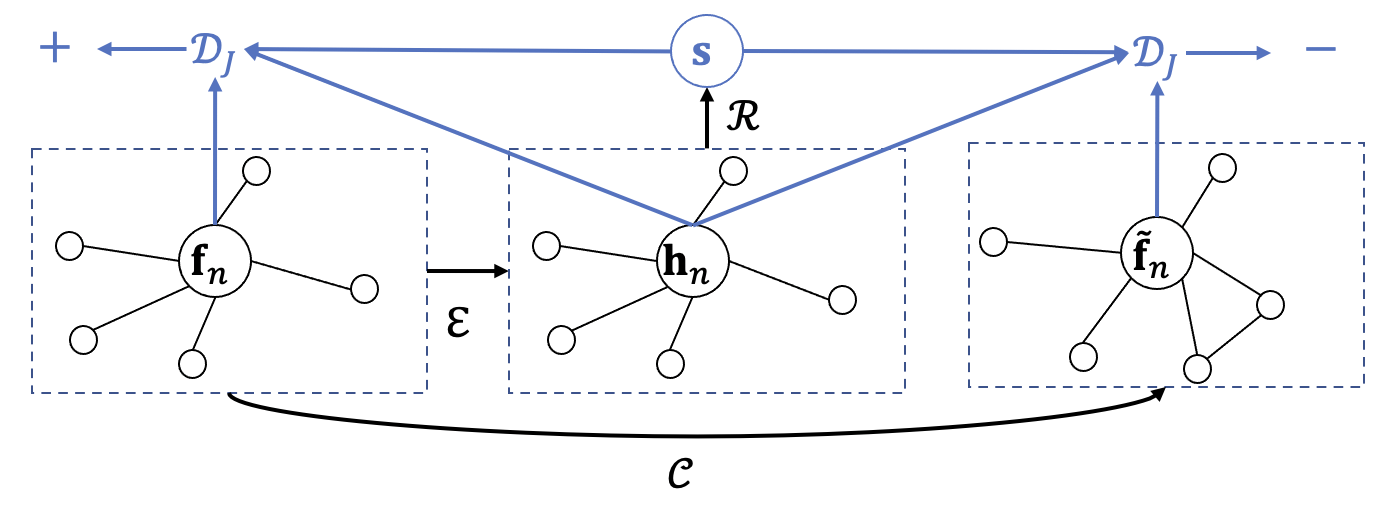}
    \caption{Illustration of the joint supervision.}\label{fig:model_joint}
\end{subfigure}
\caption{Overview of the proposed HDI. 
    The green part shows the extrinsic supervision which maximizes $I(\mathbf{h}_n; \mathbf{s})$.
    The orange part shows the intrinsic supervision which maximizes $I(\mathbf{h}_n; \mathbf{f}_n)$. 
    The blue part illustrates the joint supervision which maximizes $I(\mathbf{h}_n; \mathbf{f}_n,\mathbf{s})$.
    Detailed description is presented in Section \ref{sec:hdi}.}
    \label{fig:hdi}
\end{figure*}

\subsection{High-order Deep Infomax on Attributed Networks}\label{sec:hdi}
In this section, we propose a novel High-order Deep Infomax (HDI) for self-supervised learning on attributed networks based on DGI \cite{velivckovic2018deep} and high-order mutual information (Definition \ref{def:hmi}).
As mentioned in the introduction and preliminary (Section \ref{sec:dgi}), DGI merely considers the \textit{extrinsic} supervision signal: the mutual information between the node embedding $\mathbf{h}_n$ and the global summary vector $\mathbf{s}$ of a network.
\textit{Intrinsic} signal, i.e., the mutual dependence between node embedding $\mathbf{h}_n$ and attributes $\mathbf{f}_n$ has not been fully explored.
We introduce high-order mutual information to simultaneously capture the extrinsic and intrinsic supervision signals as well as the synergy between them in Section \ref{sec:hdi_est}.
We describe the details of HDI in Section \ref{sec:hdi_model}.

\subsubsection{High-order Mutual Information Estimation}\label{sec:hdi_est}
For the $n$-th node in the attributed network $\mathcal{G}$, we propose to jointly maximize the mutual information for three random variables: node embedding $\mathbf{h}_n$, the summary vector $\mathbf{s}$ and node attributes $\mathbf{f}_n$. 

According to the definition of the high-order mutual information (Definition \ref{def:hmi}), when $N=3$, we have:
\begin{equation}\label{eq:n=3}
\begin{split}
    I(X_1; X_2; X_3) &= H(X_1) + H(X_2) + H(X_3)\\
    &- H(X_1,X_2) - H(X_1,X_3) - H(X_2,X_3)\\
    &+ H(X_1,X_2,X_3)
\end{split}
\end{equation}
The above equation can be re-written as: 
\begin{equation}
\begin{split}
    I(X_1; X_2; X_3)
    =& H(X_1) + H(X_2) - H(X_1, X_2)\\
    +& H(X_1) + H(X_3) - H(X_1, X_3)\\
    -& H(X_1) - H(X_2, X_3) + H(X_1, X_2, X_3)\\
    =& I(X_1; X_2) + I(X_1; X_3) - I(X_1; X_2,X_3) \\
\end{split}
\end{equation}
where $I(X_1; X_2,X_3)$ denotes the mutual information between the distribution of $X_1$ with the joint distribution of $X_2$ and $X_3$.

Therefore, we have:
\begin{equation}\label{eq:interaction_information}
    I(\mathbf{h}_n; \mathbf{s}; \mathbf{f}_n) = I(\mathbf{h}_n; \mathbf{s}) + I(\mathbf{h}_n; \mathbf{f}_n) - I(\mathbf{h}_n; \mathbf{s}, \mathbf{f}_n)
\end{equation}

In the above equation, $I(\mathbf{h}_n; \mathbf{s})$ captures the extrinsic supervision signal: the mutual dependence between node embedding and the global summary.
$I(\mathbf{h}_n; \mathbf{f}_n)$ captures the intrinsic supervision signal: the mutual dependence between node embedding and attributes.
$I(\mathbf{h}_n; \mathbf{s},\mathbf{f}_n)$ 
captures the interaction between the extrinsic and intrinsic signals.

Existing methods \cite{belghazi2018mine, mukherjee2020ccmi} are unable to directly handle the proposed high-order mutual information since they are designed for mutual information of two random variables.
Therefore, following the difference-based estimation \cite{mukherjee2020ccmi}, we could estimate the high-order mutual information by estimating each of the mutual information on the right-hand side of Equation \eqref{eq:interaction_information} separately, namely $I(\mathbf{h}_n; \mathbf{s})$, $I(\mathbf{h}_n; \mathbf{f}_n)$, $I(\mathbf{h}_n; \mathbf{s}, \mathbf{f}_n)$.
By combining them together, we will have the following equation:
\begin{equation}\label{eq:eq7}
    \hat{I}(\mathbf{h}_n; \mathbf{s}; \mathbf{f}_n) = \hat{I}(\mathbf{h}_n; \mathbf{s}) + \hat{I}(\mathbf{h}_n; \mathbf{f}_n) - \hat{I}(\mathbf{h}_n; \mathbf{s}, \mathbf{f}_n)
\end{equation}
where $\hat{I}$ denotes the estimated mutual information.

Instead of directly maximizing the high-order mutual information, we maximize the three mutual information on the right-hand side of Equation \eqref{eq:interaction_information}-\eqref{eq:eq7}.
This is because if we directly maximize the high-order mutual information $I(\mathbf{h}_n; \mathbf{s}; \mathbf{f}_n)$, then we must minimize the joint mutual information $I(\mathbf{h}_n; \mathbf{s}, \mathbf{f}_n)$.
However, the experimental results show that maximizing the joint mutual information $I(\mathbf{h}_n; \mathbf{s}, \mathbf{f}_n)$ could also improve model's performance.
Therefore, we have the following objective:
\begin{equation}
    \mathcal{L} = \lambda_EI(\mathbf{h}_n; \mathbf{s}) + \lambda_II(\mathbf{h}_n; \mathbf{f}_n) + \lambda_JI(\mathbf{h}_n; \mathbf{s},\mathbf{f}_n)
\end{equation}
where $\lambda_E$, $\lambda_I$ and $\lambda_J$ are tunable coefficients. 


\paragraph{Extrinsic Signal}
For maximizing the extrinsic mutual dependence $I(\mathbf{h}_n; \mathbf{s})$, we follow \cite{velivckovic2018deep} and we will have: 
\begin{equation}\label{eq:extrinsic}
    \mathcal{L}_{E} = \mathbb{E}[\log\mathcal{D}_E(\mathbf{h}_n, \mathbf{s})] + \mathbb{E}[\log(1 - \mathcal{D}_E(\Tilde{\mathbf{h}}_n, \mathbf{s}))]
\end{equation}
where $\mathbf{h}_n$ and $\Tilde{\mathbf{h}}_n$ are the node embedding from the positive (i.e., original) $\mathcal{G}$ and the negative network $\Tilde{\mathcal{G}}$ respectively;
the negative network is obtained by corrupting the positive network via the corruption function $\mathcal{C}$:  $\Tilde{\mathcal{G}}$ = $\mathcal{C}(\mathcal{G})$;
$\mathcal{D}_E$ denotes the discriminator for distinguishing $\mathbf{h}_n$ and $\Tilde{\mathbf{h}}_n$.

The green part in Figure \ref{fig:model_ex_in} provides an illustration for the extrinsic signal.

\paragraph{Intrinsic Signal}
For the maximization of the intrinsic mutual dependence $I(\mathbf{h}_n; \mathbf{f}_n)$, we replace $\mathbf{s}$ with $\mathbf{f}_n$ in Equation \eqref{eq:extrinsic} and we will have:
\begin{equation}\label{eq:intrinsic}
    \mathcal{L}_{I} =\mathbb{E}[\log\mathcal{D}_I(\mathbf{h}_n, \mathbf{f}_n)] + \mathbb{E}[\log(1 - \mathcal{D}_I(\Tilde{\mathbf{h}}_n, \mathbf{f}_n))]
\end{equation}
where $\mathbf{f}_n$ denotes the attribute vector of the $n$-th node and $\mathcal{D}_I$ denotes the discriminator.

The orange part in Figure \ref{fig:model_ex_in} provides an illustration for the intrinsic signal.

\paragraph{Joint Signal}
The joint signal $I(\mathbf{h}_n; \mathbf{s},\mathbf{f}_n)$ is comprised of three random variables: $\mathbf{h}_n$, $\mathbf{s}$ and $\mathbf{f}_n$.
Rather than substituting $\mathbf{s}$ with ($\mathbf{s}, \mathbf{f}_n$) in Equation \ref{eq:extrinsic} and using the negative node embedding $\Tilde{\mathbf{h}}_n$ to build the negative sample pairs, we propose to use the negative node attributes $\Tilde{\mathbf{f}}_n$ to construct the negative samples.
This is because 
the extrinsic signal has already captured the mutual dependence between $\mathbf{h}_n$ and $\mathbf{s}$, 
as well as the independence between $\Tilde{\mathbf{h}}_n$ and $\mathbf{s}$ by maximizing Equation \ref{eq:extrinsic}.
The intrinsic signal has captured the mutual dependence between $\mathbf{h}_n$ and $\mathbf{f}_n$, as well as the independence between $\Tilde{\mathbf{h}}_n$ and $\mathbf{f}_n$ via Equation \ref{eq:intrinsic}.
If we add a new discriminator to distinguish $\mathbf{h}_n$ from $\Tilde{\mathbf{h}}_n$ given $(\mathbf{s}, \mathbf{f}_n)$, it will not bring substantial new information.
Instead, we propose to let the discriminator to distinguish $(\mathbf{h}_n, \mathbf{s},\mathbf{f}_n)$ with $(\mathbf{h}_n, \mathbf{s},\Tilde{\mathbf{f}}_n)$.
By introducing negative samples of attributes $\Tilde{\mathbf{f}}_n$, the encoder $\mathcal{E}$ can better capture the joint mutual dependence among $\mathbf{h}_n$, $\mathbf{s}$ and $\mathbf{f}_n$, especially the mutual dependence between $\mathbf{s}$ and $\mathbf{f}_n$, which is not captured by either extrinsic signal or intrinsic signal. 
Therefore, we have the following objective:
\begin{equation}\label{eq:joint}
    \mathcal{L}_{J} = \mathbb{E}[\log\mathcal{D}_J(\mathbf{h}_n, \mathbf{s},\mathbf{f}_n)] + \mathbb{E}[\log(1 - \mathcal{D}_J(\mathbf{h}_n, \mathbf{s},\Tilde{\mathbf{f}}_n))]
\end{equation}
where $\mathcal{D}_J$ denotes the discriminator for the joint signal.

The blue part in Figure \ref{fig:model_joint} provides an illustration for the joint signal.

\paragraph{Training Objective}
The final supervision signal is to maximize the following objective:
\begin{equation}\label{eq:hdi_obj}
    \mathcal{L} = \lambda_E\mathcal{L}_E + \lambda_I\mathcal{L}_I + \lambda_J\mathcal{L}_J
\end{equation}
where $\lambda_E$, $\lambda_I$ and $\lambda_J$ are tunable coefficients.

\subsubsection{Model Architecture}\label{sec:hdi_model}
We elaborates the details of the model architecture in this section.
\paragraph{Network Encoder $\mathcal{E}$}
We use a single layer GCN \cite{kipf2016semi} as $\mathcal{E}$:

\begin{equation}
    H = ReLU(\hat{D}^{-\frac{1}{2}}\hat{A}\hat{D}^{-\frac{1}{2}}FW)
\end{equation}
where $\hat{A}=A+wI\in\mathbb{R}^{N\times N}$, $w\in\mathbb{R}$ is the weight of self-connection, $\hat{D}[i,i]=\sum_{j}\hat{A}[i,j]$;
$F\in\mathbb{R}^{N\times d_F}$ and $H\in\mathbb{R}^{N\times d}$ denote attribute matrix and node embedding matrix,
$W\in\mathbb{R}^{d_F\times d}$ denotes the transition matrix,
$ReLU$ denotes the rectified linear unit activation function.

\paragraph{Readout Function $\mathcal{R}$}
We use average pooling as $\mathcal{R}$:
\begin{equation}\label{eq:read_out}
    \mathbf{s} = \mathcal{R}(H) =  \frac{1}{N}\sum_{n=1}^N\mathbf{h}_n
\end{equation}
where $\mathbf{h}_n\in\mathbb{R}^{d}$ is the $n$-th row of $H$, $\mathbf{s}\in\mathbb{R}^{d}$ is the global summary vector.

\paragraph{Dicriminator $\mathcal{D}$}
For $\mathcal{D}_E$ and  $\mathcal{D}_I$, we follow DGI \cite{velivckovic2018deep}:
\begin{align}
    \mathcal{D}_E(\mathbf{h}_n, \mathbf{s}) &= \sigma(\mathbf{h}_n^TM_E\mathbf{s})\\
    \mathcal{D}_I(\mathbf{h}_n, \mathbf{f}_n) &= \sigma(\mathbf{h}_n^TM_I\mathbf{f}_n)
\end{align}
where $M_E\in\mathbb{R}^{d\times d}$ and $M_I\in\mathbb{R}^{d\times d_F}$ are parameter matrices, 
$\sigma$ is the sigmoid activation function.

As for the discriminator for the joint signal $\mathcal{D}_J$, since $\mathbf{f}_n$ and $\mathbf{s}$ are not in the same space, we first project them into the same hidden space, and then use a bi-linear function to obtain the final scores.
\begin{align}
    \mathbf{z}_{f_n} &= \sigma(W_f\mathbf{f}_n)\\
    \mathbf{z}_s &= \sigma(W_s\mathbf{s})\\
    \mathbf{z} &= \sigma(W_z[\mathbf{z}_{f_n};\mathbf{z}_s])\\
    \mathcal{D}_J &= \sigma(\mathbf{h}_n^TM_J\mathbf{z})
\end{align}
where $W_f\in\mathbb{R}^{d\times d_F}$, $W_s\in\mathbb{R}^{d\times d}$, $W_z\in\mathbb{R}^{d\times 2d}$ and $M_J\in\mathbb{R}^{d\times d}$ are parameter matrices, 
$\sigma$ denotes the sigmoid activation function and $[;]$ denotes concatenation operation.

\paragraph{Corruption Function $\mathcal{C}$}
We follow DGI \cite{velivckovic2018deep} and use the random permutation of nodes as the corruption function.
Specifically, we randomly shuffle the rows of attribute matrix $F$. 

\subsection{High-order Deep Multiplex Infomax}\label{sec:hdmi}
In this section, we extend HDI to the multiplex network and propose a High-order Deep Multiplex Infomax (HDMI) model.
An attributed multiplex network is comprised of multiple layers of attributed networks.
In order to learn node embedding for the multiplex networks, we first learn node embedding on each of its layers separately and then combine them via a fusion module (Section \ref{sec:hdmi_fusion}). 
We leverage the high-order mutual information to train the fusion module (Section \ref{sec:hdmi_train}).

\begin{figure}
    \centering
    \includegraphics[width=.3\textwidth]{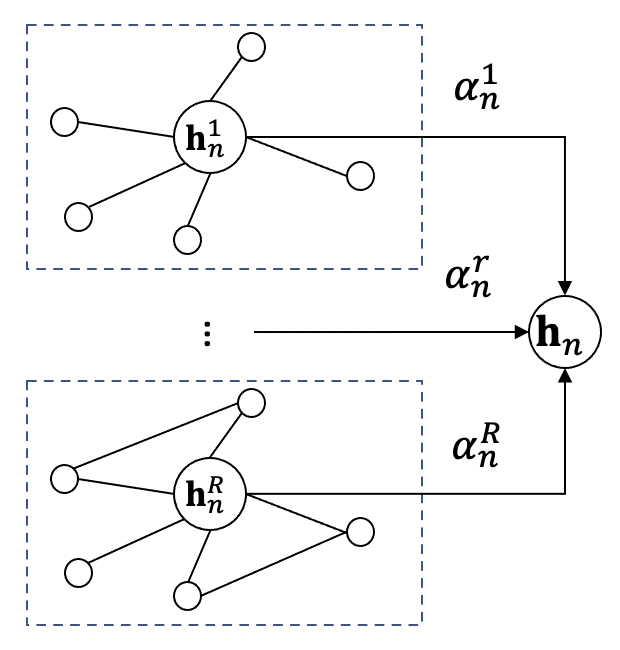}
    \caption{Illustration of the fusion module.}
    \label{fig:fusion}
\end{figure}
\subsubsection{Fusion of Embedding}\label{sec:hdmi_fusion}
The simplest way of combining node embedding from different layers is average pooling.
However, different layers are related to each other.
Therefore, we use the semantic attention \cite{you2016image} based method to fuse the node embedding, as shown in Figure \ref{fig:fusion}.

Given a node embedding from the $r$-th layer 
$\mathbf{h}_n^{r}\in\mathbb{R}^{d}$, we use layer-dependent semantic attention to obtain its score $\alpha_n^{r}$ by:
\begin{equation}
    \alpha_n^{r} = \text{tanh}( {\mathbf{y}^{r}}^TV_r\mathbf{h}_n^{r})
\end{equation}
where $V_r\in\mathbb{R}^{d'\times d}$ is the parameter matrix, $\mathbf{y}^{r}$ denotes the hidden representation vector of the $r$-th layer,
$\text{tanh}$ denotes the tangent activation function.

Then the weights of node embedding from different layers are given by:
\begin{equation}
    \alpha_n^{r} = \frac{\exp(\alpha_n^{r})}{\sum_{r'=1}^R\exp(\alpha_n^{r'})}
\end{equation}
where $R$ is the number of layers in the multiplex network.

The final embedding of the $n$-th node can be obtained by:
\begin{equation}
    \mathbf{h}_n = \sum_{r=1}^R\alpha_n^{r}\mathbf{h}_n^{r}
\end{equation}

\subsubsection{Training}\label{sec:hdmi_train}
To train the fusion module, we propose to use high-order mutual information introduced in Section \ref{sec:hdi}.
Given the fused embedding $\mathbf{h}_n$, we maximize the mutual information between $\mathbf{h}_n$ with its attributes $\mathbf{f}_n$ and global summary vector $\mathbf{s}$ of the multiplex network.
The training objective and model architecture are the same as those introduced in Section \ref{sec:hdi}, but there are two differences: the global summary vector $\mathbf{s}$ and negative node embedding $\Tilde{\mathbf{h}}_n$.
For $\mathbf{s}$ of the multiplex network, we use average pooling over the fused embedding $\mathbf{h}_n$ to obtain it.
Additionally, we obtain the negative node embedding $\Tilde{\mathbf{h}}_n$ of the multiplex network by combining the negative node embedding $\Tilde{\mathbf{h}}_n^{r}$ from different layers via the fusion module.

The fusion module can be trained jointly with HDI on different layers, and the final objective of HDMI is:
\begin{equation}\label{eq:hdmi_objective}
    \mathcal{L} = \lambda_M\mathcal{L}_M + \sum_{r=1}^R\lambda_r\mathcal{L}_r
\end{equation}
where $\mathcal{L}_M$ and $\mathcal{L}_r$ denote the objective for the fusion module and the $r$-th layer respectively, $\lambda_M$ and $\lambda_r$ are tunable coefficients. 
Note that both $\mathcal{L}_M$ and $\mathcal{L}_r$ are given by Equation \eqref{eq:hdi_obj}.

\begin{table*}[t!]
    \centering
    \caption{Statistics of the datasets}
    \begin{tabular}{c|c|c|c|c|c|c}
    \hline
    Datasets &  \# Nodes & Relation Types & \# Edges & \# Attributes & \# Labeled Data & \# Classes\\
    \hline
    \hline
    \multirow{2}{*}{ACM} & \multirow{2}{*}{3,025} & Paper-Subject-Paper (PSP) & 2,210,761 & 1,830 & \multirow{2}{*}{600} & \multirow{2}{*}{3}\\
    & & Paper-Author-Paper (PAP) & 29,281 & (Paper Abstract) & & \\
    \hline
    \multirow{2}{*}{IMDB} & \multirow{2}{*}{3,550} & Movie-Actor-Movie (MAM) & 66,428 & 1,007 & \multirow{2}{*}{300} & \multirow{2}{*}{3}\\
    & & Movie-Director-Movie (MDM) & 13,788 & (Movie plot) & \\
    \hline
    \multirow{3}{*}{DBLP} & \multirow{3}{*}{7,907} & Paper-Author-Paper (PAP) & 144,783 & \multirow{3}{*}{\shortstack{2,000 \\ (Paper Abstract)}} & \multirow{3}{*}{80} & \multirow{3}{*}{4}\\
    & & Paper-Paper-Paper (PPP) & 90,145 & & \\
    & & Paper-Author-Term-Author-Paper (PATAP) & 57,137,515 &  & \\
    \hline
    \multirow{3}{*}{Amazon} & \multirow{3}{*}{7,621} & Item-AlsoView-Item (IVI) & 266,237 & \multirow{3}{*}{\shortstack{2,000 \\ (Item description)}} & \multirow{3}{*}{80} & \multirow{3}{*}{4}\\
    & & Item-AlsoBought-Item (IBI) & 1,104,257 & & \\
    & & Item-BoughtTogether-Item (IOI) & 16,305 &  & \\
    \hline
    \end{tabular}
    \label{tab:datasets}
\end{table*}

\section{Experiments} \label{sec:experiments}
We presents the experiments to answer the following questions:
\begin{enumerate}
    \item[Q1] How will the HDI and HDMI improve the quality of the learned node embedding?
    \item[Q2] Will the fusion module assign proper attention scores to different layers?
\end{enumerate}

\subsection{Experimental Setup}
\subsubsection{Datasets}
We use the same datasets as \cite{park2020unsupervised}.
\paragraph{ACM}
The ACM\footnote{\url{https://www.acm.org/}} dataset contains 3,025 papers with two types of paper relations: paper-author-paper and paper-subject-paper.
The attribute of each paper is a 1,830-dimensional bag-of-words representation of the abstract.
The nodes are categorized into three classes: Database, Wireless Communication and Data Mining.
When training the classifier, 600 nodes are used as training samples.

\paragraph{IMDB}
The IMDB\footnote{\url{https://www.imdb.com/}} dataset contains 3,550 movies with two types of relations, including movie-actor-movie and movie-director-movie.
The attribute of each movie is a 1,007-dimensional bag-of-words representation of its plots.
The nodes are annotated with Action, Comedy or Drama, and 300 nodes are used for training classifiers.

\paragraph{DBLP}
The DBLP\footnote{\url{https://aminer.org/AMinerNetwork}} dataset \cite{tang2008arnetminer} is a multiplex network of 7,907 papers.
There are three types of relations: paper-paper, paper-author-paper, paper-author-term-author-paper.
The attribute of each paper is a 2,000-dimensional bag-of-words representation of its abstracts.
The nodes can be categorized into four categories: Data Mining, Artificial Intelligence, Computer Vision and Natural Language Processing.
80 nodes are used for training classifiers.

\paragraph{Amazon}
The Amazon\footnote{\url{https://www.amazon.com/}} dataset \cite{he2016ups} contains 7,621 items from four categories (Beauty, Automotive, Patio Lawn and Garden, and Baby) with three types of relations (Also View, Also Bought and Bought Together).
The attribute of each item is a 2,000-dimensional bag-of-words representation of its description.

\subsubsection{Comparison Methods}
We compare our proposed method with two sets of baseline methods: network embedding methods and multiplex network embedding methods.

\paragraph{Network Embedding}
\begin{itemize}
    \item Methods disregarding attributes: 
    DeepWalk \cite{perozzi2014deepwalk} and node2vec \cite{grover2016node2vec}. 
    These two methods are random-walk and skip-gram based embedding models.
    \item Methods considering attributes: GCN \cite{kipf2016semi} and GAT \cite{velivckovic2017graph}. 
    These two methods learn node embedding based on the local structure of the nodes, and the best performance of the two methods are reported. 
    DGI \cite{velivckovic2018deep} maximizes the mutual information between node embedding and the global summary vector.
    ANRL \cite{zhang2018anrl} uses skip-gram to model the local contextual topology and uses an auto-encoder to capture attribute information.
    CAN \cite{meng2019co} learns node embedding and attribute embedding in the same semantic space.
    DGCN \cite{zhuang2018dual} considers both local and global consistency. 
\end{itemize}

\paragraph{Multiplex Network Embedding}
\begin{itemize}
    \item Methods disregarding attributes:
    CMNA \cite{chu2019cross} uses the cross-network information to refine both of the inter-network and intra-network node embedding.
    MNE \cite{zhang2018scalable} uses a common embedding with multiple additional embedding from different layers for each node.
    \item Methods considering attributes:
    mGCN \cite{ma2019multi} and HAN \cite{wang2019heterogeneous} use GCN/GAT to extract node embedding for each layer of the multiplex networks and then combine them via attention mechanism.
    DMGI and DMGI$_{\text{attn}}$ \cite{park2020unsupervised} extend DGI onto multiplex networks and uses consensus regularization to combine node embedding from different layers, where DMGI$_{\text{attn}}$ leverages attention to obtain the reference node embedding.
\end{itemize}

\subsubsection{Evaluation Metrics}
We evaluate our proposed HDMI and comparison methods on both of the unsupervised tasks (i.e., clustering and similarity search \cite{wang2019heterogeneous, park2020unsupervised}) and a supervised task (i.e., node classification) with the following evaluation metrics.
\begin{itemize}
    \item Macro-F1 and Micro-F1 are used to evaluate models on the node classification task.
    \item Normalized Mutual Information (NMI) is adopted for the node clustering task.
    \item Sim@5 evaluates how well does a model project nodes with the same label to nearby positions in the embedding space. 
\end{itemize}
For the clustering and classification, we first train models with their self-supervision signals and then run the clustering methods (i.e., K-means) and classification methods (i.e., logistic regression) to obtain the NMI and Macro-F1/Micro-F1 scores.
For similarity search, we follow \cite{wang2019heterogeneous, park2020unsupervised} to first compute the cosine similarity between each pair of nodes based on their embedding.
Then for each node, the top-5 most similar nodes are selected to calculate the ratio of nodes sharing the same label (Sim@5).

\subsubsection{Implementation Details}
We set the dimension of node embedding as 128, and use Adam optimizer \cite{kingma2014adam} with the learning rate of 0.001 to optimize the models.
The weight of the self-connection for GCN is fixed as $3$.
Following \cite{park2020unsupervised}, for HDMI, we use the same discriminator for different layers.
We use grid search to tune the coefficients and report the best results.
Early stopping with a patience of 100 is adopted to prevent overfitting.

\begin{table*}[t]
    \centering
    \caption{Overall performance on the supervised task: node classification.}
    \begin{tabular}{l|cc|cc|cc|cc}
         \hline
         Dataset & \multicolumn{2}{c|}{ACM} & \multicolumn{2}{c|}{IMDB} & \multicolumn{2}{c|}{DBLP} & \multicolumn{2}{c}{Amazon} \\
         \hline
         Metric & Macro-F1 & Micro-F1 & Macro-F1 & Micro-F1 & Macro-F1 & Micro-F1 & Macro-F1 & Micro-F1\\
         \hline
         DeepWalk & 0.739 & 0.748 & 0.532 & 0.550 & 0.533 & 0.537  & 0.663 & 0.671 \\
         node2vec & 0.741 & 0.749 & 0.533 & 0.550 & 0.543 & 0.547 & 0.662 & 0.669\\
         GCN/GAT & 0.869 & 0.870 & 0.603 & 0.611 & 0.734 & 0.717 & 0.646 & 0.649\\
         DGI & 0.881 & 0.881 & 0.598 & 0.606 & 0.723 & 0.720 & 0.403 & 0.418\\
         ANRL & 0.819 & 0.820 & 0.573 & 0.576 & 0.770 & 0.699 & 0.692 & 0.690\\
         CAN & 0.590 & 0.636 & 0.577 & 0.588 & 0.702 & 0.694 & 0.498 & 0.499\\
         DGCN & 0.888 & 0.888 & 0.582 & 0.592 & 0.707&  0.698 & 0.478 & 0.509\\
         \hline
         CMNA & 0.782 & 0.788 & 0.549 & 0.566 & 0.566 & 0.561 & 0.657 & 0.665\\
         MNE & 0.792 & 0.797 & 0.552 & 0.574 & 0.566 & 0.562 & 0.556 & 0.567\\
         mGCN & 0.858 & 0.860 & 0.623 & 0.630 & 0.725 & 0.713 & 0.660 & 0.661\\
         HAN & 0.878 & 0.879 & 0.599 & 0.607 & 0.716 & 0.708 & 0.501 & 0.509\\
         DMGI & 0.898 & 0.898 & 0.648 & 0.648 & 0.771 & 0.766 & 0.746 & 0.748\\
         DMGI$_{\text{attn}}$ & 0.887 & 0.887 & 0.602 & 0.606 & 0.778 & 0.770 & 0.758 & 0.758\\
         \hline
         HDI & \textbf{0.901} & 0.900 & 0.634 & 0.638 & 0.814 & 0.800 & 0.804 & 0.806 \\
         HDMI & \textbf{0.901} & \textbf{0.901} & \textbf{0.650} & \textbf{0.658} & \textbf{0.820} & \textbf{0.811} & \textbf{0.808} & \textbf{0.812}\\
         \hline
    \end{tabular}
    \label{tab:overall_classification}
\end{table*}

\begin{table*}[t]
    \centering
    \caption{Overall performance on the unsupervised tasks: node clustering and similarity search.
    }
    \begin{tabular}{l|cc|cc|cc|cc}
         \hline
         Dataset & \multicolumn{2}{c|}{ACM} & \multicolumn{2}{c|}{IMDB} & \multicolumn{2}{c|}{DBLP} & \multicolumn{2}{c}{Amazon} \\
         \hline
         Metric & NMI & Sim@5 & NMI & Sim@5 & NMI & Sim@5 & NMI & Sim@5\\
         \hline
         DeepWalk & 0.310 & 0.710 & 0.117 & 0.490 & 0.348 & 0.629  & 0.083 & 0.726 \\
         node2vec & 0.309 & 0.710 & 0.123 & 0.487 & 0.382 & 0.629 & 0.074 & 0.738\\
         GCN/GAT & 0.671 & 0.867 & 0.176 & 0.565 & 0.465 & 0.724 & 0.287 & 0.624\\
         DGI & 0.640 & 0.889 & 0.182 & 0.578 & 0.551 & 0.786 & 0.007 & 0.558\\
         ANRL & 0.515 & 0.814 & 0.163 & 0.527 & 0.332 & 0.720 & 0.166 & 0.763\\
         CAN & 0.504 & 0.836 & 0.074 & 0.544 & 0.323 & 0.792 & 0.001 & 0.537\\
         DGCN & 0.691 & 0.690 & 0.143 & 0.179 & 0.462 & 0.491 & 0.143 & 0.194\\
         \hline
         CMNA & 0.498 & 0.363 & 0.152 & 0.069 & 0.420 & 0.511 & 0.070 & 0.435\\
         MNE & 0.545 & 0.791 & 0.013 & 0.482 & 0.136 & 0.711 & 0.001 & 0.395\\
         mGCN & 0.668 & 0.873 & 0.183 & 0.550 & 0.468 & 0.726 & 0.301 & 0.630\\
         HAN & 0.658 & 0.872 & 0.164 & 0.561 & 0.472 & 0.779 & 0.029 & 0.495\\
         DMGI & 0.687 & 0.898 & 0.196 & 0.605 & 0.409 & 0.766 & 0.425 & 0.816\\
         DMGI$_{\text{attn}}$ & \textbf{0.702} & \textbf{0.901} & 0.185 & 0.586 & 0.554 & 0.798 & 0.412 & 0.825\\
         \hline
         HDI & 0.650 & 0.900 & 0.194 & 0.605 & 0.570 & 0.799 & 0.487 & 0.856 \\
         HDMI & 0.695 & 0.898 & \textbf{0.198} & \textbf{0.607} & \textbf{0.582} & \textbf{0.809} & \textbf{0.500} & \textbf{0.857}\\
         \hline
    \end{tabular}
    \label{tab:overall_clustering}
\end{table*}

\subsection{Quantitative Evaluation}
In this section, we present the experimental results on both supervised and unsupervised downstream tasks to quantitatively demonstrate the effectiveness of the proposed supervision signals as well as the proposed models HDI and HDMI.

\subsubsection{Overall Performance}
We present the experimental results of node classification, node clustering and similarity search on the multiplex networks in Table \ref{tab:overall_classification} and Table \ref{tab:overall_clustering}.
HDI separately learns embedding on different layers and uses average pooling to combine node embedding from different layers.
Table \ref{tab:overall_classification} shows that HDMI outperforms the state-of-the-art models for all of the supervised tasks, and Table \ref{tab:overall_clustering} shows that HDMI achieves higher scores for most of the cases on the unsupervised tasks.
These results demonstrate that the embedding extracted by HDMI is more discriminative.
Additionally, the scores of HDI are generally competitive to the state-of-the-art methods, which demonstrate the effectiveness of the proposed HDI to a certain degree.

\subsubsection{Ablation Study}
\paragraph{Performance of the fusion module}
To evaluate the proposed fusion module, we compare it with average pooling.
As shown in Table \ref{tab:overall_classification} and Table \ref{tab:overall_clustering}, HDMI outperforms HDI on all metrics for all datasets, except for the Sim@5 on the ACM dataset. 
However, it can be noted that the gap is very tiny.

\paragraph{Performance of different supervision signals}
We compare the performance of the Extrinsic (E.), Intrinsic (I.), and Joint (J) signal as well as the Reconstruction (R.) error on attributed networks (different layers of the multiplex networks) in the datasets, and present experimental results in Table \ref{tab:ablation_classification} and Table \ref{tab:ablation_clustering}.
Firstly, incorporating the mutual dependence between embedding and attributes could improve the model's performance on both supervised and unsupervised tasks, which can be observed by comparing E. with E.+R. and E.+I. 
Secondly, maximizing mutual information between node embedding and attributes (E.+I.) is better than minimizing the reconstruction error of attributes given node embedding (E.+R.).
Thirdly, the joint signal (E.+I.+J.) can further improve the discriminative ability of node embedding for the node classification task.
Finally, combining node embedding from different layers (lower parts of Table \ref{tab:ablation_classification}-\ref{tab:ablation_clustering}) will result in better results, indicating that different layers can mutually help each other.

\subsection{Qualitative Evaluation}
In this section, we present the qualitative evaluation results.

\subsubsection{Visualization of node embedding}
\begin{figure*}
    \centering
    \begin{subfigure}[b]{.245\textwidth}
        \includegraphics[width=1\textwidth]{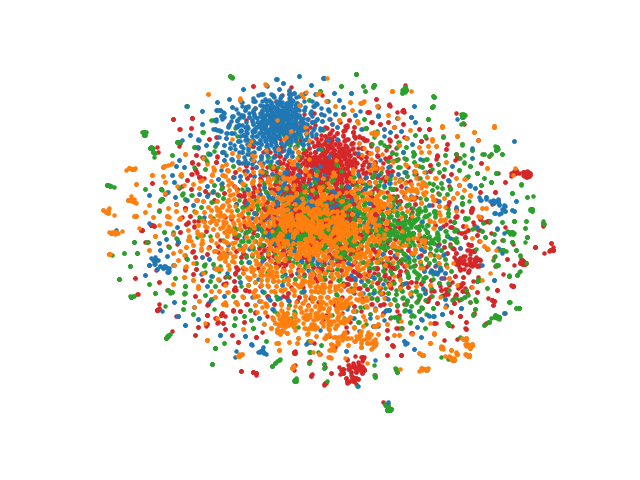}
        \caption{E.}
    \end{subfigure}
    \begin{subfigure}[b]{.245\textwidth}
        \includegraphics[width=1\textwidth]{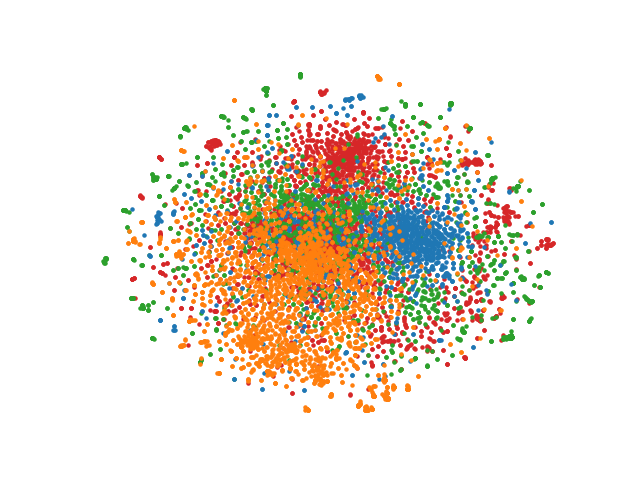}
        \caption{E. + R.}\label{fig:visualize_re}
    \end{subfigure}
    \begin{subfigure}[b]{.245\textwidth}
        \includegraphics[width=1\textwidth]{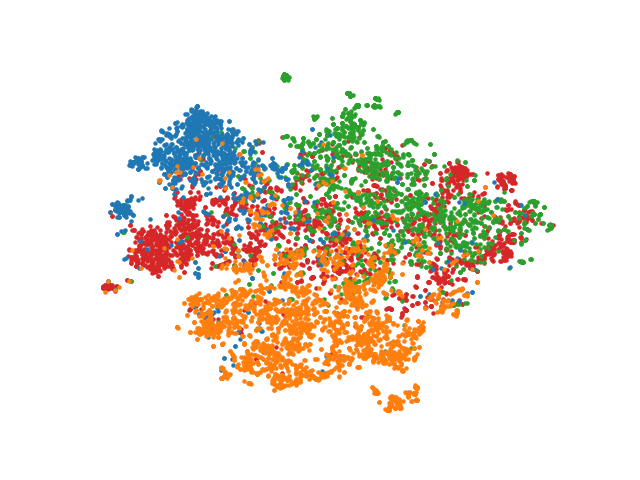}
        \caption{E. + I.}\label{fig:visualize_ei}
    \end{subfigure}
    \begin{subfigure}[b]{.245\textwidth}\label{fig:visualize_eeij}
        \includegraphics[width=1\textwidth]{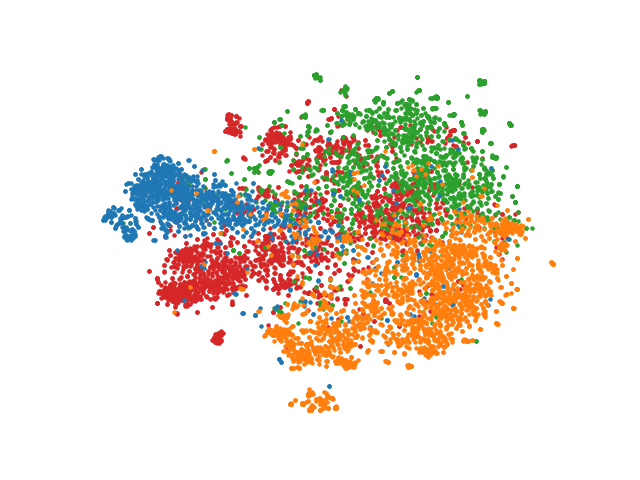}
        \caption{E. + I. + J.}\label{fig:visualize_eij}
    \end{subfigure}
    \caption{Visualization of the node embedding on the IOI layer of the Amazon dataset. The four different colors represent four different classes of the nodes. E., I., J., and R. denote the Extrinsic, Intrinsic, Joint mutual information and Reconstruction error.}
    \label{fig:visualization}
\end{figure*}

\paragraph{Node embedding of an attributed network}
We show the t-sne \cite{maaten2008visualizing} visualization of the node embedding of the IOI layer in the Amazon dataset learned by the Extrinsic (E.), the Extrinsic + Reconstruction (E. + R.), the Extrinsic + Intrinsic (E. + I.) and the Extrinsic + Intrinsic + Joint (E. + I. + J.) signals in Figure \ref{fig:visualization}.
Different colors in the figure represent different classes.
It is obvious that the intrinsic signal and the joint signal significantly improve the discriminative ability of the node embedding, and the joint signal can further improve the quality of the node embedding
Additionally, Figure \ref{fig:visualize_re} shows that the reconstruction error does not substantially connect with the discriminative ability of the node embedding.

\begin{figure*}
    \centering
    \begin{subfigure}[b]{.245\textwidth}
        \includegraphics[width=1\textwidth]{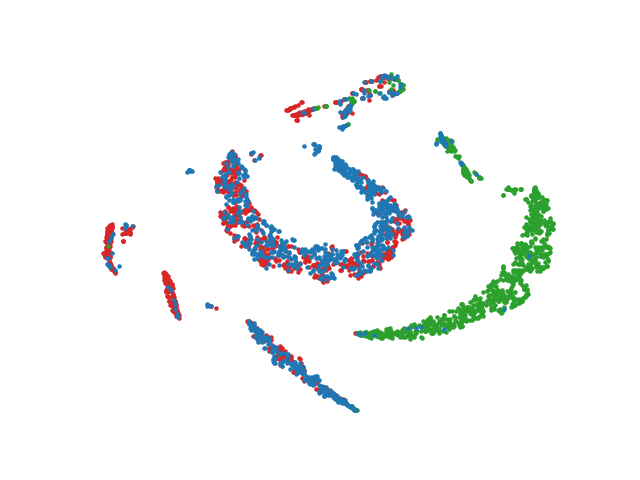}
        \caption{PSP}\label{fig:visualize_psp}
    \end{subfigure}
    \begin{subfigure}[b]{.245\textwidth}
        \includegraphics[width=1\textwidth]{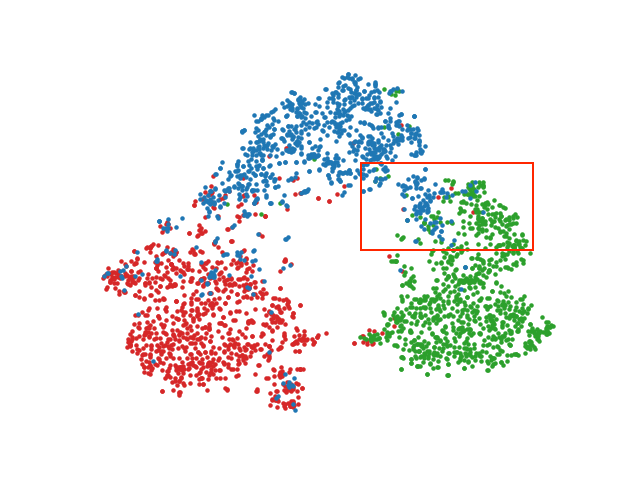}
        \caption{PAP}\label{fig:visualize_pap}
    \end{subfigure}
    \begin{subfigure}[b]{.245\textwidth}
        \includegraphics[width=1\textwidth]{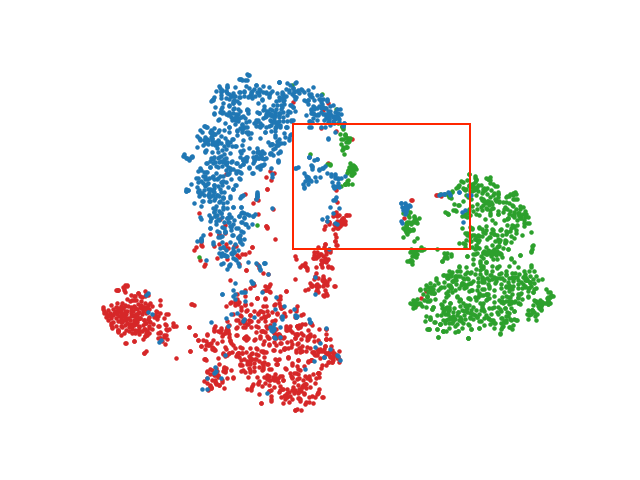}
        \caption{Average}\label{fig:visualize_avg}
    \end{subfigure}
    \begin{subfigure}[b]{.245\textwidth}
        \includegraphics[width=1\textwidth]{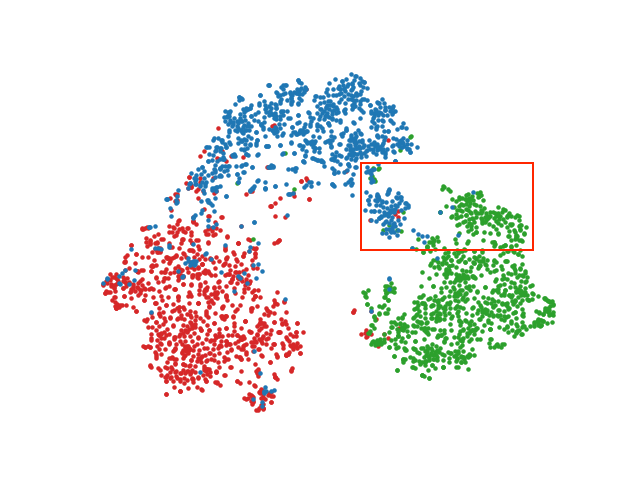}
        \caption{Fusion}\label{fig:visualize_fusion}
    \end{subfigure}
    \caption{Visualization of the node embedding on the ACM dataset. PSP and PAP are two layers of the ACM multiplex network. Different colors represent different classes of the nodes.
    Average and fusion denote using average pooling and the proposed fusion module to obtain the combined embedding.
    As shown in the red boxes, the fusion module better separates different classes than average pooling.}
    \label{fig:visualization2}
\end{figure*}

\paragraph{Node embedding of a multiplex network}
The t-sne visualization for the node embedding learned by HDI on the PSP and the PAP layer of the ACM multiplex network are presented in Figure \ref{fig:visualize_psp}-\ref{fig:visualize_pap}. 
Figure \ref{fig:visualize_avg}-\ref{fig:visualize_fusion} show the combined embedding obtained by average pooling and fusion module. 
As can be noted in the red boxes of Figure \ref{fig:visualize_pap}-\ref{fig:visualize_avg}, average pooling for the embedding of different layers better separates nodes than the embedding learned from PAP layer alone.
Additionally, Figure \ref{fig:visualize_avg}-\ref{fig:visualize_fusion} show that the fusion module separates the nodes even better than the average pooling.

\subsubsection{Attention scores}
\begin{figure*}
    \centering
    \begin{subfigure}[b]{.2\textwidth}
        \includegraphics[width=1\textwidth]{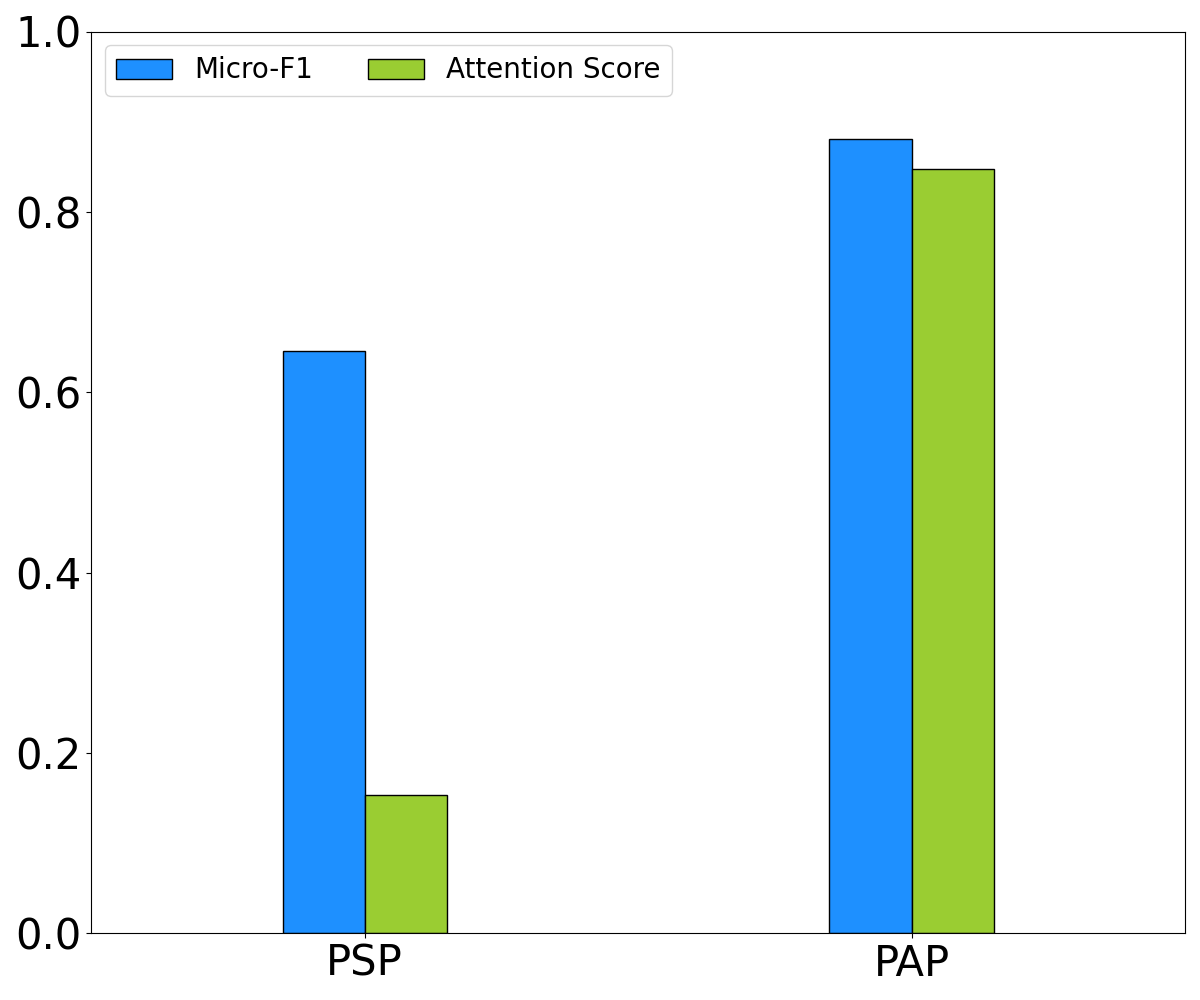}
        \caption{ACM}
    \end{subfigure}\quad\quad\,
    \begin{subfigure}[b]{.2\textwidth}
        \includegraphics[width=1\textwidth]{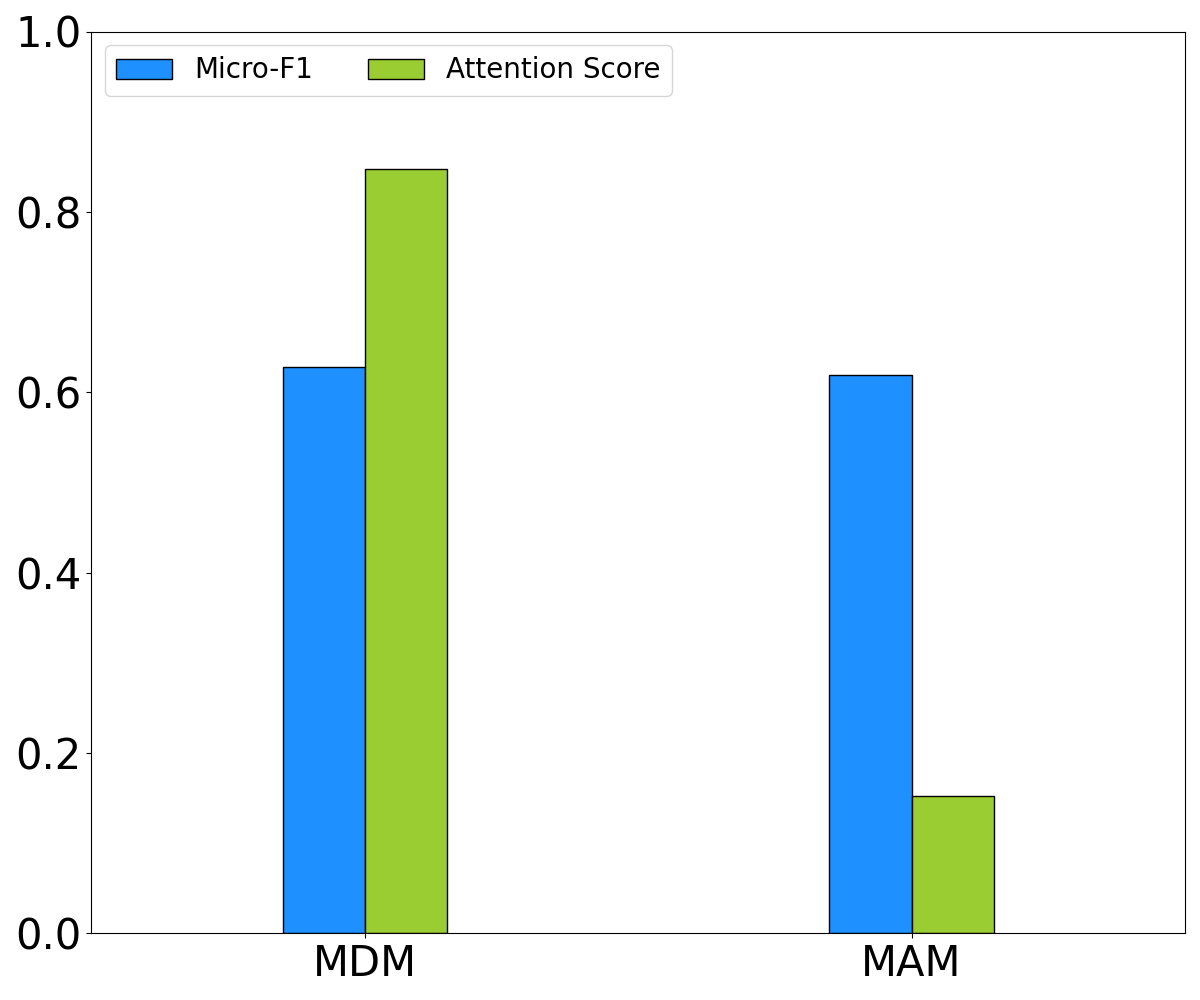}
        \caption{IMDB}
    \end{subfigure}\quad\quad\,
    \begin{subfigure}[b]{.2\textwidth}
        \includegraphics[width=1\textwidth]{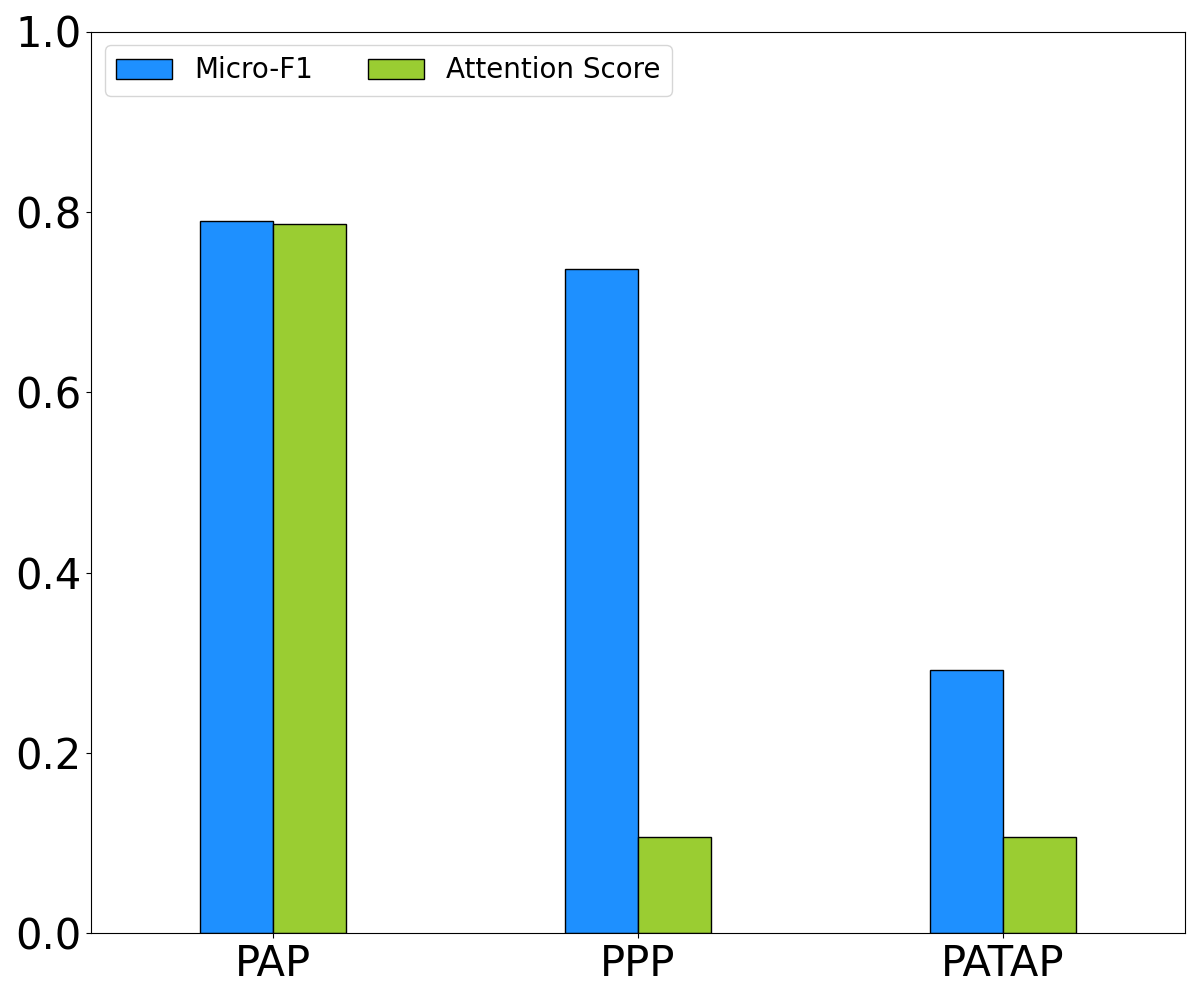}
        \caption{DBLP}
    \end{subfigure}\quad\quad\,
    \begin{subfigure}[b]{.2\textwidth}
        \includegraphics[width=1\textwidth]{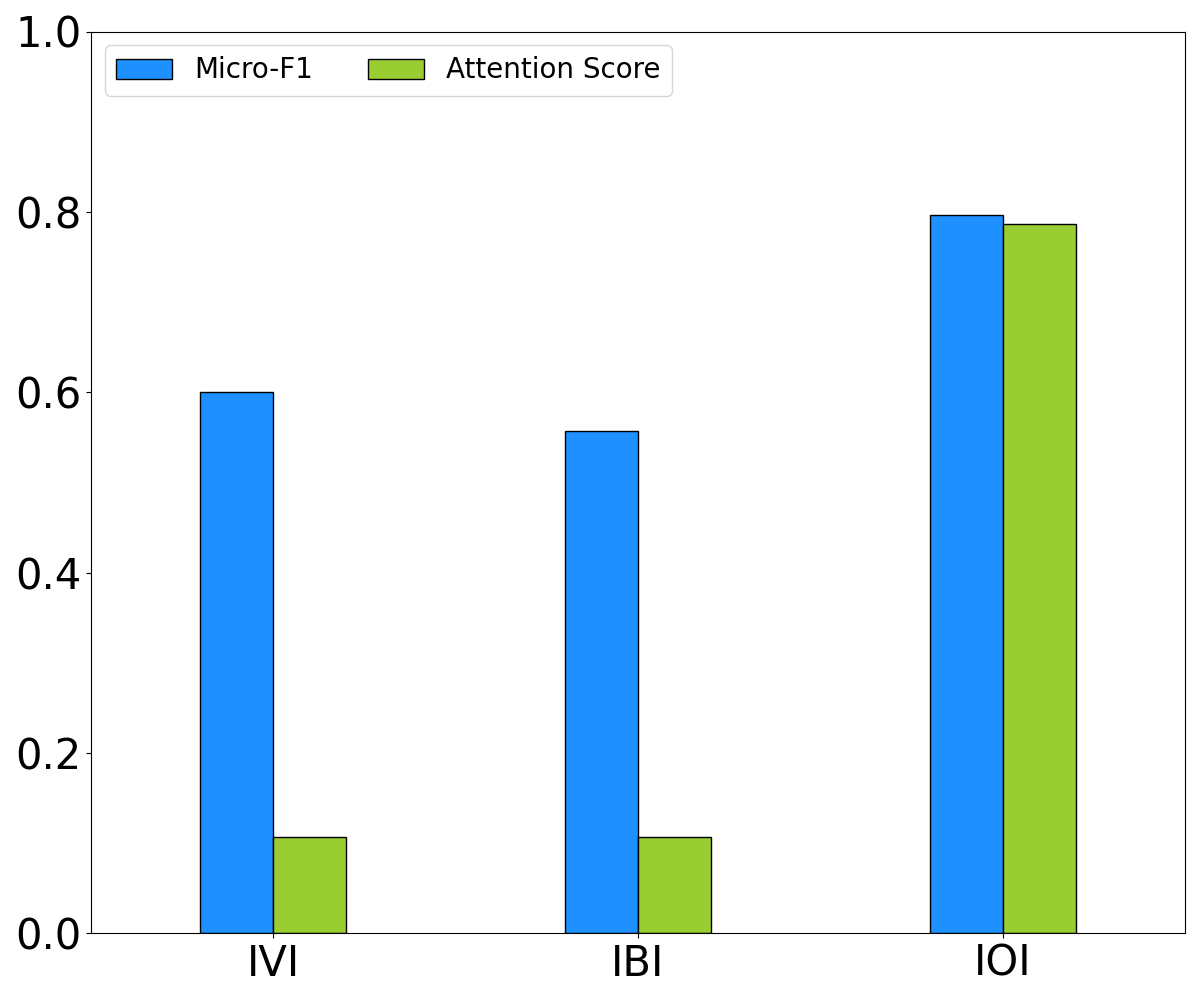}
        \caption{Amazon}
    \end{subfigure}
    \caption{Attention scores and Micro-F1 scores for each layer.}
    \label{fig:att}
\end{figure*}

We present the visualization of attention scores and the Micro-F1 scores for each layer of the multiplex networks in Figure \ref{fig:att}.
Generally, for the layers that have higher Micro-F1 scores, their attention scores are also higher, demonstrating the effectiveness of the proposed fusion module.

\begin{table*}[t]
    \scriptsize
    \centering
    \caption{Ablation study on the supervised task: node classification. MaF1 and MiF1 denote Macro-F1 and Micro-F1. E., I., J., and R. denote the Extrinsic, Intrinsic, Joint mutual information and Reconstruction error. The results of HDMI$_{\text{avg}}$ and HDMI are copied from Table \ref{tab:overall_classification} to better show the effectiveness of combining different layers.}
    \begin{tabular}{l|cc|cc|cc|cc|cc|cc|cc|cc|cc|cc}
        \hline
        Dataset & \multicolumn{4}{c|}{ACM} & \multicolumn{4}{c|}{IMDB} & \multicolumn{6}{c|}{DBLP} & \multicolumn{6}{c}{Amazon} \\
        \hline
        Layer & \multicolumn{2}{c|}{PSP} & \multicolumn{2}{c|}{PAP} & \multicolumn{2}{c|}{MDM} & \multicolumn{2}{c|}{MAM} & \multicolumn{2}{c|}{PAP} & \multicolumn{2}{c|}{PPP} & \multicolumn{2}{c|}{PATAP} & \multicolumn{2}{c|}{IVI} & \multicolumn{2}{c|}{IBI} & \multicolumn{2}{c}{IOI}\\
        \hline
        Metric & MaF1 & MiF1 & MaF1 & MiF1 & MaF1 & MiF1 & MaF1 & MiF1 & MaF1 & MiF1 & MaF1 & MiF1 & MaF1 & MiF1 & MaF1 & MiF1 & MaF1 & MiF1 & MaF1 & MiF1\\
        \hline
        E. & 0.663 & 0.668 & 0.855 & 0.853 & 0.573 & 0.586 & 0.558 & 0.564 & 0.804 & 0.796 & 0.728 & 0.717 & 0.240 & 0.272 & 0.380 & 0.388 & 0.386 & 0.410 & 0.569 & 0.574\\
        E. + R. & 0.668 & 0.673 & 0.864 & 0.847 & 0.590 & 0.597 & 0.560 & 0.570 &
        0.809 & 0.801 & 0.737 & 0.728 & 0.240 & 0.280 & 0.392 & 0.398 & 0.410 & 0.427 & 0.579 & 0.589\\
        \hline
        E. + I. & 0.719 & 0.732 & 0.886 & 0.887 & 0.617 & 0.624 & 0.593 & 0.600 & 0.803 & 0.792 & 0.742 & 0.733 & 0.240 & 0.276 & 0.559 & 0.561 & 0.517 & 0.527 & 0.792 & \textbf{0.799}\\
        E. + I. + J. & \textbf{0.742} & \textbf{0.744} & \textbf{0.889} & \textbf{0.888} & \textbf{0.626} & \textbf{0.631} & \textbf{0.600} & \textbf{0.606} & \textbf{0.812} & \textbf{0.803} & \textbf{0.751} & \textbf{0.745} & \textbf{0.241} & \textbf{0.284} & \textbf{0.581} & \textbf{0.583} &  \textbf{0.524} & \textbf{0.529} & \textbf{0.796} & \textbf{0.799}\\
        \hline
        \hline
         Metric & \multicolumn{2}{c}{MaF1} & \multicolumn{2}{c|}{MiF1} & \multicolumn{2}{c}{MaF1} & \multicolumn{2}{c|}{MiF1} & \multicolumn{3}{c}{MaF1} & \multicolumn{3}{c|}{FiF1} & \multicolumn{3}{c}{MaF1} & \multicolumn{3}{c}{MiF1}\\
        \hline
        HDI & \multicolumn{2}{c}{\textbf{0.901}} & \multicolumn{2}{c|}{0.900} & \multicolumn{2}{c}{0.634} & \multicolumn{2}{c|}{0.638} & \multicolumn{3}{c}{0.814} & \multicolumn{3}{c|}{0.800} & \multicolumn{3}{c}{0.804} & \multicolumn{3}{c}{0.806} \\
        HDMI & \multicolumn{2}{c}{\textbf{0.901}} & \multicolumn{2}{c|}{\textbf{0.901}} & \multicolumn{2}{c}{\textbf{0.650}} & \multicolumn{2}{c|}{\textbf{0.658}} & \multicolumn{3}{c}{\textbf{0.820}} & \multicolumn{3}{c|}{\textbf{0.811}} & \multicolumn{3}{c}{\textbf{0.808}} & \multicolumn{3}{c}{\textbf{0.812}}\\
        \hline
    \end{tabular}
    \label{tab:ablation_classification}
\end{table*}

\begin{table*}[t]
    \scriptsize
    \centering
    \caption{Ablation study on the unsupervised tasks: node clustering and similarity search. S@5 denotes Sim@5. E., I., J., and R. denote the Extrinsic, Intrinsic, Joint mutual information and Reconstruction error.
    The results of HDMI$_{\text{avg}}$ and HDMI are copied from Table \ref{tab:overall_clustering} to better show the effectiveness of combining different layers.}
    \begin{tabular}{l|cc|cc|cc|cc|cc|cc|cc|cc|cc|cc}
        \hline
        Dataset & \multicolumn{4}{c|}{ACM} & \multicolumn{4}{c|}{IMDB} & \multicolumn{6}{c|}{DBLP} & \multicolumn{6}{c}{Amazon} \\
        \hline
        Layer & \multicolumn{2}{c|}{PSP} & \multicolumn{2}{c|}{PAP} & \multicolumn{2}{c|}{MDM} & \multicolumn{2}{c|}{MAM} & \multicolumn{2}{c|}{PAP} & \multicolumn{2}{c|}{PPP} & \multicolumn{2}{c|}{PATAP} & \multicolumn{2}{c|}{IVI} & \multicolumn{2}{c|}{IBI} & \multicolumn{2}{c}{IOI}\\
        \hline
        Metric & NMI & S@5 & NMI & S@5 & NMI & S@5 & NMI & S@5 & NMI & S@5 & NMI & S@5 & NMI & S@5 & NMI & S@5 & NMI & S@5 & NMI & S@5\\
        \hline
        E. & 0.526 & 0.698 & 0.651 & 0.872 & 0.145 & 0.549 & 0.089 & 0.495 & 0.547 & 0.800 & 0.404 & 0.741 & 0.054 & 0.583 & 0.002 & 0.395 & 0.003 & 0.414 & 0.038 & 0.701\\
        E. + R. & 0.525 & \textbf{0.728} & 0.659 & 0.874 & 0.150 & 0.552 & 0.079 & 0.490 & 0.564 & 0.804 & \textbf{0.421} & 0.741 & 0.051 & 0.568 & 0.002 & 0.399 & 0.003 & 0.426 & 0.020 & 0.660 \\
        \hline
        E. + I. & 0.527 & 0.708 & 0.656 & 0.882 & 0.193 & \textbf{0.595} & \textbf{0.143} & \textbf{0.527} & \textbf{0.569} & 0.802 & 0.405 & 0.741 & 0.053 & 0.569 & 0.152 & 0.512 & 0.143 & 0.517 & 0.401 & 0.824\\
        E. + I. + J. & \textbf{0.528} & 0.716 & \textbf{0.662} & \textbf{0.886} & \textbf{0.194} & 0.592 & \textbf{0.143} & \textbf{0.527} & 0.562 & \textbf{0.805} & 0.408 & \textbf{0.742} & \textbf{0.054} & \textbf{0.591} & \textbf{0.169} & \textbf{0.544} & \textbf{0.153} & \textbf{0.525} & \textbf{0.407} & \textbf{0.826}\\
        \hline
        \hline
         Metric & \multicolumn{2}{c}{NMI} & \multicolumn{2}{c|}{Sim@5} & \multicolumn{2}{c}{NMI} & \multicolumn{2}{c|}{Sim@5} & \multicolumn{3}{c}{NMI} & \multicolumn{3}{c|}{Sim@5} & \multicolumn{3}{c}{NMI} & \multicolumn{3}{c}{Sim@5}\\
        \hline
        HDI & \multicolumn{2}{c}{0.650} & \multicolumn{2}{c|}{\textbf{0.900}} & \multicolumn{2}{c}{0.194} & \multicolumn{2}{c|}{0.605} & \multicolumn{3}{c}{0.570} & \multicolumn{3}{c|}{0.799} & \multicolumn{3}{c}{0.487} & \multicolumn{3}{c}{0.856} \\
        HDMI & \multicolumn{2}{c}{\textbf{0.695}} & \multicolumn{2}{c|}{0.898} & \multicolumn{2}{c}{\textbf{0.198}} & \multicolumn{2}{c|}{\textbf{0.607}} & \multicolumn{3}{c}{\textbf{0.582}} & \multicolumn{3}{c|}{\textbf{0.809}} & \multicolumn{3}{c}{\textbf{0.500}} & \multicolumn{3}{c}{\textbf{0.857}}\\
        \hline
    \end{tabular}
    \label{tab:ablation_clustering}
\end{table*}

\section{Related Works}\label{sec:related_works}
In this section, we briefly review the most relevant works, including the self-supervised graph representation learning (Section \ref{sec:related1}), the graph neural networks (Section \ref{sec:related2}), as well as the mutual information methods (Section \ref{sec:related3}).

\subsection{Self-supervised Network Representation Learning}\label{sec:related1}
In this section, we mainly focus on self-supervised network embedding methods.
For a comprehensive review, please refer to \cite{cui2018survey, zhang2019graph, zhou2018graph}.
We leave the mutual information related methods to Section \ref{sec:related2}, and graph neural networks with other signals to Section \ref{sec:related3}.

Inspired by word2vec \cite{mikolov2013distributed}, deepwalk \cite{perozzi2014deepwalk} and node2vec \cite{grover2016node2vec} leverage random walks to sample context of a node, and then learn node embedding by maximizing the probabilities of the sampled neighbors.
SDNE \cite{wang2016structural} and LINE \cite{tang2015line} propose to learn node embedding based on the first-order and the second-order proximities between nodes.
Struct2vec \cite{wang2016structural} further designs a training signal based on the local structural similarity of the nodes.
However, these methods ignore node attributes.
Zhang et al. \cite{zhang2018anrl} uses a neighbor enhancement auto-encoder to encoder attributes with the re-construction error.
Meng et al. \cite{meng2019co} propose to co-embed attributes and nodes into the same semantic space via Variational Auto-Encoder (VAE) \cite{kingma2013auto}, but VAE tends to \textit{minimize} the mutual information between inputs and hidden features \cite{zhao2017infovae}.

The multiplex network is also known as multi-dimensional network \cite{ma2018multi}. 
Liu et al. \cite{liu2017principled} propose a random walk based method to map multiplex networks into a vector space.
Zhang et al. \cite{zhang2018scalable} propose a scalable multiplex network embedding method trained via negative sampling.
Shi et al. \cite{shi2018mvn2vec} leverage preservation and collaboration to learn node embedding.
Ma et al. \cite{ma2019multi} propose mGCN for multiplex networks trained via negative sampling.
Cen et al. \cite{cen2019representation} use the random walk based method to train a unified network for attributed multiplex networks.
Ban et al. \cite{yikun2019no} introduce a density metric as the training signal to improve the performance of node clustering.
Li et al. \cite{li2018multi} introduce MANE considering both within-layer and cross-layer dependency.

\subsection{Graph Neural Networks}\label{sec:related2}
In this section, we briefly review the graph neural networks with external training signals (e.g. cross-entropy between predicted labels and real labels).
For a comprehensive review of the graph neural networks, please refer to \cite{zhou2018graph, zhang2019graph}. 

Kipf et al. \cite{kipf2016semi} propose Graph Convolutional Network (GCN) based on \cite{defferrard2016convolutional}.
Hamilton et al. \cite{hamilton2017inductive} propose an aggregation-based model called GraphSAGE. 
Graph Attention Network (GAT) \cite{velivckovic2017graph} learns different weights for a node's neighbors by attention mechanism.
Zhuang et al. \cite{zhuang2018dual} propose DGCN which considers both local and global consistency.

Qu et al. \cite{qu2017attention} use an attention mechanism to embed multiplex networks into a single collaborated embedding.
Wang et al. \cite{wang2019heterogeneous} propose an attention based method called HAN for merging node embedding from different layers.
Chu et al. \cite{chu2019cross} propose CrossMNA which leverages cross-network information for network alignment.
Yan et al. \cite{yan2021dynamic} introduce DINGAL for dynamic knowledge graph alignment. 
Jing et al.\cite{jing2021network} propose TGCN for modeling high-order tensor graphs.

\subsection{Mutual Information Methods}\label{sec:related3}
Belghazi et al. \cite{belghazi2018mine} propose MINE to estimate mutual information of two random variables by neural networks.
Mukherjee et al. \cite{mukherjee2020ccmi} propose a classifier based neural estimator for conditional mutual information.
Recently, the infomax principle \cite{linsker1988self} has been used for self-supervised representation learning in computer vision \cite{hjelm2018learning} and speech recognition \cite{ravanelli2018learning} by maximizing the mutual information between different hidden features.
In the field of network representation learning, DGI \cite{velivckovic2018deep} maximizes the mutual information of node embedding with the global summary vector.
Peng et al. \cite{peng2020graph} propose GMI for homogeneous networks.
Park et al. \cite{park2020unsupervised, park2020deep} extend DGI onto multiplex networks and propose a consensus regularization to combine embedding of different layers.

\section{Conclusion}\label{sec:conclusion}
In this paper, we introduce a novel High-order Deep Multiplex Infomax (HDMI) to learn network embedding for multiplex networks via self-supervised learning.
Firstly, we propose a High-order Deep Infomax (HDI) and use high-order mutual information as the training signal for attributed networks. 
The proposed signal simultaneously captures the extrinsic signal (i.e., the mutual dependence between node embedding and the global summary), the intrinsic signal (i.e., the mutual dependence between node embedding and attributes), and the interaction between these two signals.
Secondly, we propose a fusion module based on the attention mechanism to combine node embedding from different layers.
We evaluate the proposed HDMI on four real-world datasets for both unsupervised and supervised downstream tasks.
The results demonstrate the effectiveness of the proposed HDMI, HDI, and the fusion module.


\begin{acks}
This work is supported by National Science Foundation under grant No. 1947135, by the NSF Program on Fairness in AI in collaboration with Amazon under award No. 1939725, and by the National Research Foundation of Korea (NRF) grant funded by the Korea government (MSIT) (No. 2021R1C1C1009081) 
The content of the information in this document does not necessarily reflect the position or the policy of the Government or Amazon, and no official endorsement should be inferred.  The U.S. Government is authorized to reproduce and distribute reprints for Government purposes notwithstanding any copyright notation here on.
\end{acks}

\bibliographystyle{ACM-Reference-Format}
\bibliography{acmart.bib}


\end{document}